\documentclass{article}
\usepackage{ijcai22}

\usepackage{times}
\usepackage{soul}
\usepackage{url}
\usepackage[hidelinks]{hyperref}
\usepackage[utf8]{inputenc}
\usepackage[small]{caption}
\usepackage{graphicx}
\usepackage{amsmath}
\usepackage{amsthm}
\usepackage{multirow}
\usepackage{booktabs}
\usepackage{algorithm}
\usepackage{algorithmic}
\usepackage{amsfonts}
\usepackage{bbm}
\title{Learning Consistency from High-quality Pseudo-labels for Weakly Supervised Object Localization}

\author{
Kangbo Sun$^{1,2}$\and
Jie Zhu$^{1,2,}$\footnote{Corresponding Author}\\
\affiliations
$^1$Shanghai Jiao Tong University, China\\
$^2$Shanghai Frontier Science Research Center for Gravitational Wave Detection, China\\
\emails
\{kangbosun, zhujie\}@sjtu.edu.cn
}

\begin{document}

\maketitle

\begin{abstract}
	Pseudo-supervised learning methods have been shown to be effective for weakly supervised object localization tasks.
	However, the effectiveness depends on the powerful regularization ability of deep neural networks.
	Based on the assumption that the localization network should have similar location predictions on different versions of the same image, 
	we propose a two-stage approach to learn more consistent localization.
	In the first stage, we propose a mask-based pseudo label generator algorithm, 
	and use the pseudo-supervised learning method to initialize an object localization network.
	In the second stage, we propose a simple and effective method for evaluating the confidence of pseudo-labels based on classification discrimination, 
	and by learning consistency from high-quality pseudo-labels, we further refine the localization network to get better localization performance.
	Experimental results show that our proposed approach achieves excellent performance in three benchmark datasets including CUB-200-2011, ImageNet-1k and Tiny-ImageNet,
	which demonstrates its effectiveness.
\end{abstract}

\section{Introduction}

Thanks to numerous and accurate manual location labels, deep learning has achieved great success in fully supervised object localization tasks.
Recently, weakly supervised learning methods that require less manual labeling have become a hot spot.
Different from fully supervised object localization, 
Weakly Supervised Object Localization (WSOL) aims to learn to classify and localize a single object in the image with only class labels.
Without location labels, it is difficult to directly optimize deep neural networks, which is a huge challenge.

The mainstream methods \cite{selvaraju2017grad,zhang2018adversarial,choe2019attention}  are  based on Class Activation Mapping (CAM) \cite{zhou2016learning} .
Instead of directly predicting the bounding box,
CAM-based  methods mainly focus on how to use the feature map extracted by CNN to localize the image's discriminative region. 
Specifically, CAM-based methods take the high-response region in the corresponding CAM as the location of object,
which has proven to be intuitive and effective. 
However, the optimization direction of the classification network tends to 
have the largest response-value in the most discriminative region. 
The predicted location by CAM-based methods usually is the most discriminative part of the object, 
which could not cover the entire object region well.
Some methods  similar to "erase" \cite{zhang2018adversarial} or "dropout-layer" \cite{choe2019attention} have been proposed to alleviate this problem, 
however, the localization performance of CAM-based methods is still not satisfactory.

Zhang et al. \cite{zhang2020rethinking} proposed a direct method named PSOL, 
which proposed to train an additional regression network to directly predict bounding boxes under the supervision of pseudo bounding boxes generated by CAM or DDT \cite{wei2019unsupervised}.  
PSOL has proved that although the deviation between the pseudo bounding boxes and the ground-truth  bounding boxes is not negligible, 
it is still possible to train a localization network with higher localization accuracy.
However, the performance improvement of PSOL depends on the powerful regularization ability of deep neural networks. 
Intuitively, if effective  prior regularization could be applied to retain high-quality pseudo-labels and further refine the localization network, 
the localization performance could be further improved.

Consistency regularization is widely adopted to leverage the unlabeled data in semi-supervised classification tasks,
which is based on the assumption that the classification network should have similar class predictions on  different versions of the same image.
We believe that the localization network should also have similar location predictions on different versions of the same image.
In our work, we propose a two-stage approach for weakly supervised object localization tasks. 
Our approach includes the initialization stage and the refinement stage.
In the initialization stage, a mask-based pseudo bounding box generator is proposed to predict high-precision pseudo-labels to 
initialize an object localization network.
In the refinement stage, a confidence evaluation method is proposed to evaluate the quality of the prediction of localization network, 
and consistency regularization is adopted to refine the object localization network with high-confidence pseudo-labels.
We name our proposed approach as Learning Consistency from High-quality Pseudo-label (LCHP). 
We evaluate our proposed approach on three benchmark datasets including CUB-200-2011, ImageNet-1k and Tiny-ImageNet. 
Our  proposed approach  achieves state-of-the-art performance on CUB-200-2011 and Tiny-ImageNet, and achieves a comparable  performance compared with state-of-the-art methods on ImageNet-1k.

\section{Related Works}

\textbf{Weakly Supervised Object Localization.} 
Weakly supervised object localization aims to learn to classify and localize with only class labels.
It is assumpted that there is only one instance in the image, which leads the localization tasks in WSOL to become a bounding box prediction tasks. 
To localize objects without any location annotations, CAM \cite{zhou2016learning} proposed to generate class activation mapping, and determined high-response region as object's location.
Grad-CAM \cite{selvaraju2017grad} proposed to replace the feature maps with gradients to generate more accurate CAM.
CAM-based methods have the drawback of only locating the most discriminative part of the object.
To address this issue, ACoL \cite{zhang2018adversarial} proposed to erase the most discriminative features in the feature map to discover the more complete object's region.
SPG \cite{zhang2018self} proposed to generate self-produced guidance masks to localize the object.
ADL \cite{choe2019attention} proposed the dropout layer to cover the entire object.
Moreover, Zhang et al. \cite{zhang2020rethinking} proposed PSOL.
PSOL is the first method to adopt pseudo-labeling to train a regression network to directly localize  objects in WSOL.
PSOL  proposed to generate pseudo bounding boxes for the training images through a co-supervised localization method.
The co-supervised localization method in PSOL is based on Deep Descriptor Transforming (DDT) \cite{wei2019unsupervised},
which needs to perform PCA (Principal Component Analysis) on the CNN feature maps of all the training images to obtain cross-image location information.
Indeed, the initialization stage of our proposed LCHP  follows the PSOL paradigm.
However, in our work, we adopt a simple mask-based pseudo-label generation algorithm, which 
does not require  the across-image location information.

\textbf{Consistency Regularization.} 
Consistency regularization is widely adopted in semi-supervised learning methods.
UDA \cite{xie2019unsupervised}, ReMixMatch \cite{berthelot2019remixmatch} and FixMatch \cite{sohn2020fixmatch} all predict pseudo-labels on  weakly-augmented examples 
and enforce consistency against with the predictions of strongly-augmented examples. 
In our work, the refinement stage of our proposed LCHP adopts the FixMatch-way to refine the localization network.
However, unlike FixMatch for classification tasks, we have designed a novel pseudo-label confidence evaluation method  and image augmentation specifically for object localization tasks.

\section{Approach}

\begin{figure*}[h]
	\centering
	\includegraphics[scale=0.55]{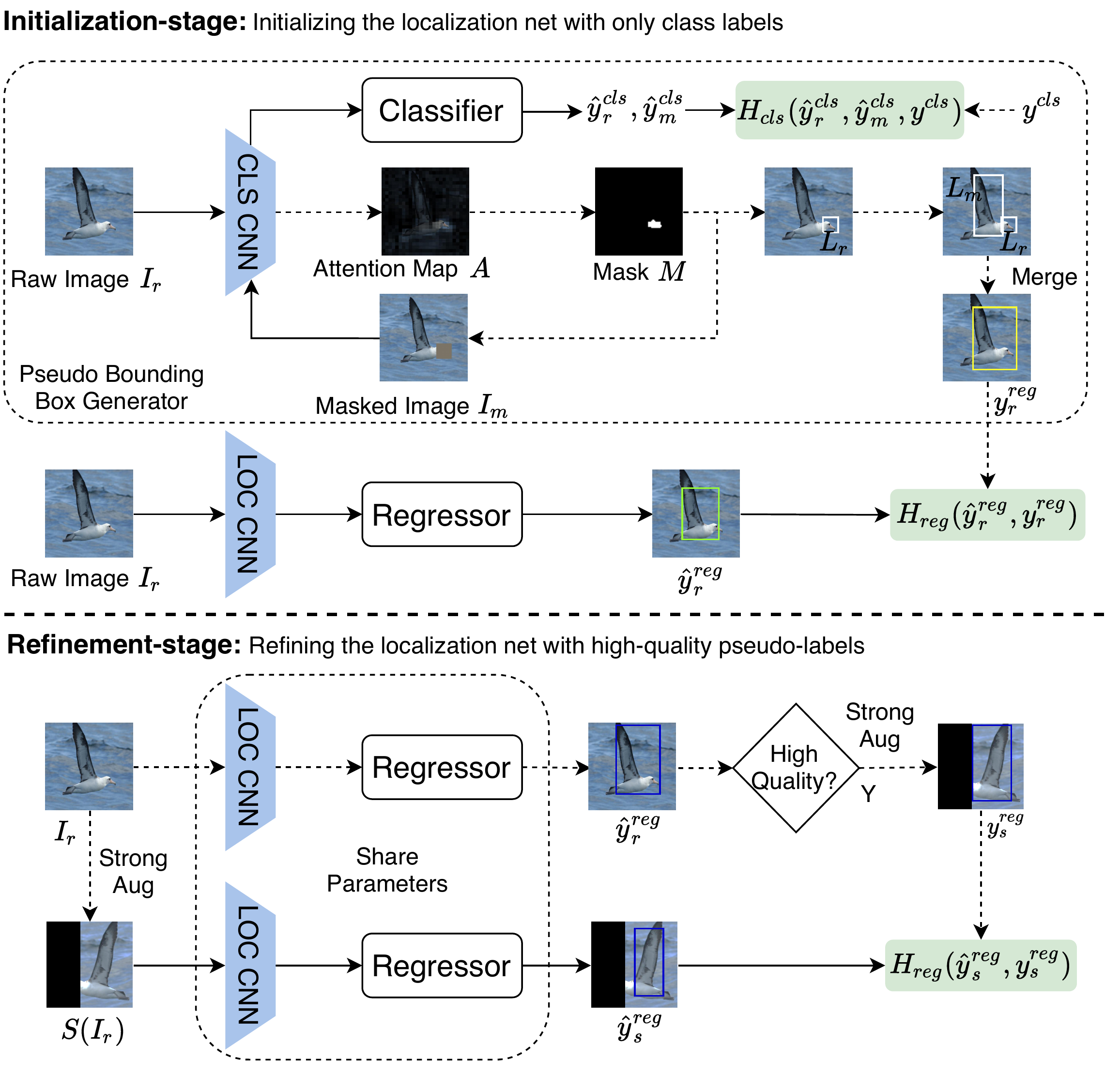}
	\caption{
		The overview of our proposed LCHP, which is a two-stage approach including the initialization and refinement stage.
        In the initialization stage, we generate pseudo bounding boxes with the supervision of class labels, and further train an initialized localization network.
        In the refinement stage, we evaluate the quality of the predicted bounding boxes, and retain the high-quality samples to refine the localization with consistency regularization.
	}
	\label{fig:overview}
	\vspace*{-3mm}
\end{figure*}

In this section, we first introduce how we train an initialized localization network with only class labels in the initialization stage.
Finally, we introduce how to further refine the object localization network with high-quality pseudo bounding boxes in the refinement stage.

\subsection{Initialization}

In our work, we adopt a simple mask-based pseudo-label generation algorithm
that uses the CNN feature maps as the clue to localize the foreground object. 
Specifically, we average the feature maps on channel-wise to get attention map $A$, and define the pixels in $A$ with a value greater than the preset threshold $\delta$ are foreground pixels, 
Then, the smallest rectangle containing all foreground pixels could be regarded as the discriminative region $L_r$.
However, the discriminative region $L_r$ obtained by classification is usually the most discriminative part of the object, which cannot cover the entire object well.
To address this issue, we adopt the method of mask-out. 
By masking the discriminative region $L_r$ in the raw image $I_r$ with the value of  zero, we could obtain the masked image $I_m$ as shown in Fig.\ref{fig:overview}.
Similarly, we could localize the most discriminative region $L_m$ in $I_m$.
By merging the two obtained discriminative regions, we can get a higher-precision pseudo bounding box.
The description of our mask-based pseudo-label generation algorithm is shown in Algorithm \ref{alg:pesudo}.

\begin{algorithm}[h]
	\caption{Our pseudo bounding box generator}
	\label{alg:pesudo}
	\begin{algorithmic}
	\REQUIRE input image $I_r$, threshold $\delta$, classification CNN
	\STATE \textbf{1.} Generate feature maps $F_r$ with CNN from $I_r$
	\STATE \textbf{2.} Generate $A_r$ by averaging $F_r$ on channels
	\STATE \textbf{3.} Binarize $A_r$ into $M$ according to the threshold $\delta$ 
	\STATE \textbf{4.} Localize the discriminative region $L_r$.
	\STATE \textbf{5.} Generate the masked image $I_m $
	\STATE \textbf{6.} Repeat process 1-4 with input image $I_m$ once, and get the discriminative region $L_m$
	\STATE \textbf{7.} Merge the local regions $L_r$ and $L_m$, and generate pesudo bounding box $y^{reg}_{r}$
	\ENSURE pesudo bounding box $y^{reg}_{r}$
	\end{algorithmic}
\end{algorithm}

To better to localize the discriminative regions, we optimize the classification network  not only on the raw images $I_r$, but also on the the masked images $I_m$.
We denote $H_{cls}(y^{cls}_1,y^{cls}_2,y^{cls}_3)$  as the average of the cross entropy between the classification probability distributions $y^{cls}_1$ and $y^{cls}_3$  
and the cross entropy between the classification probability distributions $y^{cls}_2$ and $y^{cls}_3$.
The classification loss $\mathcal{L}_{cls}$ is determined as follows:
\begin{equation}
	\begin{aligned}
		\mathcal{L}_{cls} &  = \frac{1}{N} \sum_{i}^{N} H_{cls}(\hat{y}^{cls}_{i,r},\hat{y}^{cls}_{i,m}, y^{cls}_{i})  \\
		& = - \frac{1}{2N}  \sum_{i}^{N} y^{cls}_{i} (log (\hat{y}^{cls}_{i,r})  + log (\hat{y}^{cls}_{i,m}))
	\end{aligned}
		\label{equ:clsloss}
\end{equation}
where $N$ is the size of mini-batch, $y_{i}^{cls}$ is the $i$-th one-hot class label,
$\hat{y}_{i,r}^{cls}$  and $\hat{y}_{i,m}^{cls}$ are the classification probability distributions of the $i$-th raw image $I_{i,r}$ and the $i$-th masked image $I_{i,m}$, respectively.

For the simplicity of training, we train the localization network together while training the classification network.
We denote $H_{reg}(y^{reg}_1,y^{reg}_2)$  as the mean square error between the bounding box $y^{reg}_1$ and $y^{reg}_2$.
The regression loss in the initialization stage is determined as follows:
\begin{equation}
	\begin{aligned}
		\mathcal{L}_{reg} & = \frac{1}{N} \sum_{i}^{N} H_{reg}(\hat{y}^{reg}_{i,r} ,y^{reg}_{i,r}) \\
		& = \frac{1}{N} \sum_{i}^{N} ||  \hat{y}^{reg}_{i,r} - y^{reg}_{i,r} ||^2
	\end{aligned}
		\label{equ:regloss}
\end{equation}
where $\hat{y}_{i,r}^{reg}$  and $y_{i,r}^{reg}$ are the predicted bounding box and the pseudo bounding box, respectively,

Therefore, the training loss in the initialization stage could be determined as follows:
\begin{equation}
	\begin{aligned}
		\mathcal{L}_{init} = \mathcal{L}_{cls} + \alpha \mathcal{L}_{reg}
	\end{aligned}
		\label{equ:initloss}
\end{equation}
where $\alpha$ is to balance the classification and regression loss.

\subsection{Refinement}
\label{sec:refinement}

The pseudo-supervised learning method has been shown to be effective in optimizing  localization network. 
The effectiveness comes from the powerful regularization capabilities of deep neural networks,
which could overcome the acceptable deviation between pseudo labels and ground-truth labels.
Intuitively, the smaller deviation, the better performance of the network.
For this reason, we expect to retain high-quality pseudo-labels from the generated pseudo-labels on the training dataset to further optimize the object localization network.

\textbf{High-quality Pseudo-labels.} 
To obtain high-quality pseudo-labels, we need to evaluate the confidence of the predicted bounding boxes.
In classification tasks, researchers \cite{xie2019unsupervised,berthelot2019remixmatch,sohn2020fixmatch} usually use the maximum value in the predicted classification probability distributions as the classification confidence.
In object localization tasks, intuitively, if the classification network is strong enough, the maximum classification probability value of the foreground-relevant region will be much larger than the foreground-irrelevant region.
Based on this prior knowledge, we propose a confidence evaluation method for pseudo bounding box labels based on classification discrimination.
We denote the cropped image in the bounding box from the raw image $I_r$ as $I_b$.  
We denote the maximum value of the predicted classification distributions of  $I_b$ as the confidence, and denote the bounding box with confidence greater than the preset threshold $\tau$ as high-quality pseudo-labels.
The high-quality indicator matrix can be defined as $\mathbbm{1}(max(y^{cls}_{b})> \tau)$, where $y^{cls}_{b}$ is the the classification probability distributions of $I_b$.

\textbf{Learning Consistency.}
To learn better localization performance, we adopt the FixMatch-way \cite{sohn2020fixmatch} to refine the localization network.
As shown in the refinement stage in Fig.\ref{fig:overview}, for the input raw image $I_r$, 
we use the initialized localization network to predict its  bounding box $\hat{y}^{reg}_{r} = Loc(I_r)$, 
where $Loc(\cdot)$ is the mapping function representing the localization network.
Then, stronger forms of augmentation are adopted to obtain 
the strong augmented image $S(I_r)$, where $S(\cdot)$ means the strong augmentation that can be applied on  images and  bounding boxes together.
Similarly, the strong augmented bounding box logit $\hat{y}_{s}^{reg}$
and the pseudo bounding box label $y^{reg}_{s}$ are determined as follows:
\begin{equation}
	\left\{
	\begin{aligned}
		\hat{y}_{s}^{reg} & = Loc(S(I_r)) \\
		y^{reg}_{s} &= S(\hat{y}^{reg}_{r})  =   S(Loc(I_r)) \\
	\end{aligned}
	\right. 
	\label{equ:reg_logit_label}
\end{equation}

Therefore, to refine the localization with consistency regularization, the refinement loss $\mathcal{L}_{refine}$ is determined as follows:
\begin{equation}
\begin{aligned}
	\mathcal{L}_{refine} &= \frac{1}{N} \sum_{i=1}^{N} H_{reg}(\hat{y}^{reg}_{i,s}, y^{reg}_{i,s}) \\
	&= \frac{1}{N} \sum_{i=1}^{N} ||\hat{y}_{i,s}^{reg} - y_{i,s}^{reg}||^2 \\
\end{aligned}
	\label{equ:refineloss1}
\end{equation}

Further, with only high-quality pesudo-labels, Equation (\ref{equ:refineloss1}) could be rewritten as:
\begin{equation}
	\mathcal{L}_{refine} = \frac{1}{N} \sum_{i=1}^{N} \mathbbm{1}(max(\hat{y}_{i,b}^{cls}) > \tau)  ||\hat{y}_{i,s}^{reg} - y_{i,s}^{reg}||^2
	\label{equ:refineloss2}
\end{equation}
where $\tau$ is the threshold for retaining the high-quality bounding boxes.

\section{Experiments}

\subsection{Datasets}
We evaluate our proposed approach on three benchmark datasets: CUB-200-2011 \cite{wah2011caltech}, ImageNet-1k \cite{deng2009imagenet} and Tiny-ImageNet \cite{le2015tiny}. 
CUB-200-2011 is a bird dataset with 200 classes, containing 5994 training images and 5794 testing images. 
Each image in CUB-200-2011 has been labeled with a bounding box annotation. 
ImageNet-1k is a large dataset with 1000 classes, containing 1,281,197 training images and 50,000 validation images. 
Each image in CUB-200-2011 validation has been labeled with no less than one bounding box annotation, which means that there may be more than one instance in each image.
Tiny-ImageNet is a subset of the ImageNet dataset. The dataset contains 100,000 images of 200 classes (500 for each class) with the solution of 64x64. 
Each class has 500 training images, 50 validation images, and 50 test images. 
Each image in the training and validation dataset has been labeled with an accurate bounding box. 
We train all models on the training dataset with only class labels, and evaluate models on the testing dataset of CUB-200-2011 and the validation datasets of ImageNet-1k and Tiny-ImageNet.

\subsection{Metrics}
We follow previous state-of-the-art methods \cite{zhou2016learning,choe2019attention,zhang2020rethinking} to evaluate our approach.
The metrics includes \textit{GT-Konwn} localization accuracy and \textit{Top-1/5} localization accuracy.
\textit{GT-Konwn} accuracy is the localization accuracy with known ground truth class.
\textit{GT-Konwn} is correct when the intersection over union (IoU) between the predicted bounding box and the ground truth bounding box is 50\% or more.
\textit{Top-1/5} is correct when the predicted top-1/5 class label and \textit{GT-Konwn} are both correct.

\begin{table*}[htbp]
	\centering
	\caption{The performances (\%)  comparison with state-of-the-art methods on CUB-200-2011 testing dataset and ImageNet-1k validation dataset. The best performance has been bolded, and the second best performance has been underlined.}
	\label{tab:cub-results}
	  \begin{tabular}{lccccccc}
	  \toprule
	  \multicolumn{1}{c}{\multirow{2}[2]{*}{Method}} & \multicolumn{1}{c}{\multirow{2}[2]{*}{Backbone}} & \multicolumn{3}{c}{CUB-200-2011} & \multicolumn{3}{c}{ImageNet-1k} \\
			&       & \textit{Top-1} & \textit{Top-5} & \textit{GT-Known} & \textit{Top-1} & \textit{Top-5} & \textit{GT-Known} \\
	  \midrule
	  CAM \cite{zhou2016learning} 					& GoogLeNet-GAP 		& ---   & ---   & 41.00   & 43.60   & 57.00   & --- \\
	  Grad-CAM \cite{selvaraju2017grad}				& VGG16  				& ---   & ---   & ---   & 43.49   & 53.59   & --- \\
	  ACoL  \cite{zhang2018adversarial}				& VGG-GAP 				& 45.92   & 56.51   & ---   & 45.83   & 59.43   & 62.96 \\
	  SPG \cite{zhang2018self}   					& InceptionV3 			& 46.64   & 57.72   & ---   & 48.60   & 60.00   & 64.69 \\
	  CutMix \cite{yun2019cutmix}   				& ResNet50  			& 54.81   & ---   & ---   & 47.25   & ---   & --- \\
	  ADL   \cite{choe2019attention}   				& ResNet50-SE 			& 62.29   & ---   & ---   & 48.53   & ---   & --- \\
	  PSOL \cite{zhang2020rethinking}				& DenseNet161  			& \underline{74.97}   & \underline{89.12}   & \underline{92.54}   & \textbf{55.31}   & \textbf{64.18}   & \textbf{66.28} \\
	  LayerCAM \cite{jiang2021layercam}				& VGG16  				& ---   & ---   & ---   & 47.24   & 58.74   & --- \\
	  CSoA   \cite{kou2021improve}   				& GoogLeNet 			& 62.31   & 73.51   & ---   & 51.19   & 62.54   & \underline{66.20} \\
	  \midrule
	  Pseudo Label (Our baseline)    				& InceptionV3-BAP 		& 61.03   & 70.21   & 72.44   & 47.21   & 55.37   & 57.57 \\
	  LCHP-I 										& InceptionV3 			& 73.61   & 84.60   & 87.12   & 52.59   & 61.67   & 64.11 \\
	  LCHP-R 										& InceptionV3 			& \textbf{80.39}   & \textbf{91.72}   & \textbf{94.36}   & \underline{54.12}   & \underline{63.51}   & 66.08 \\
	  \bottomrule
	  \end{tabular}
	\vspace*{-3mm}
\end{table*}

\subsection{Experimental Details}

\textbf{General Details.}
We train all models using Stochastic Gradient Descent (SGD) optimizer with a momentum of 0.9, weight decay of 1e-5, 
and the size of batch is set to 32 on one RTX 3090 GPU. 
In the initialization stage, the number of total epochs is set to 40, 40, and 10 for CUB-200-2011, Tiny-ImageNet and ImageNet-1k respectively.
$\alpha$ is set to 20 for balancing the classification and regression loss.
The  initial learning rate is set to 2e-3.
Specifically, we reduce the learning rate of the classification network with exponential decay of 0.9 after every epoch, 
while the learning rate of the localization network remains unchanged, which follows PSOL.
In the refinement stage, the number of total epochs  is set to 40, 40, and 5 for CUB-200-2011, Tiny-ImageNet and ImageNet-1k respectively.
The  initial learning rate is set to 2e-3 for CUB-200-2011, Tiny-ImageNe datasets,  2e-4 for ImageNet-1k dataset.
Without special instructions, 
the  $\delta$ is set to 0.7 for CUB-200-2011 and ImageNet-1k datasets, 0.8 for Tiny-ImageNet dataset,
and the  $\tau$ is set to 0.9 for all the three datasets.

\textbf{Classification Backbone.}
We adopt the InceptionV3-BAP \cite{hu2019see} as our classification backbone.
The backbone  extracts the output feature of layer \textit{Mix6e} from InceptionV3 model and utilizes the 1*1 convolution layer to generate attention maps, 
and finally uses the bilinear pooling \cite{lin2015bilinear} to generate bilinear features. 
The number of attention maps is determined as 32. The InceptionV3 network is pre-trained on ImageNet-1k dataset. 
We use the signed square root and L2 normalization after bilinear pooling, which is widely applied in \cite{lin2015bilinear,gao2016compact,yu2018hierarchical}.

\textbf{Momentum Update.}
The optimization direction of  the classification network based on cross-entropy loss is not consistent with that of the localization network.
We find that the distribution of the pseudo bounding box generated by our Algorithm \ref{alg:pesudo} is unstable between two adjacent iterations,
which  makes the training of the localization network unstable.
To address this issue, we adopt the momentum update method.
It is assumpted that $\theta_1$ are the parameters of the classification CNN, and $\theta_2$ are the CNN parameters used to generate pseudo bounding boxes in the initialization stage.
Then each iteration has the following Equation:
\begin{equation}
	\theta_2 = \beta \theta_2 + (1-\beta) \theta_1
	\label{equ:momentum}
\end{equation}
where $\beta$ is set to 0.9 in our models.

\textbf{Augmentation.}
We use two kinds of augmentation including  general augmentation and  strong augmentation.
The general augmentation is used to generate the input raw images $I_r$.
The strong augmentation is adopted to obtain strong perturbed versions of images in the refinement stage.
For general augmentation, we resized the input images to 512x512 and randomly cropped images into 448x448. 
Besides, we also use random horizontal flip with a probability of 50\% for general augmentation. 
For strong augmentation, we adopt the implementation by imagaug \cite{imgaug}, which is an image augmentation library that can transform images and bounding boxes together.
We adopt three kinds of augmentation including scale, translation, and flip. Details are described as follows:

$\bullet$ \textbf{Scale:} we scale images to a value of 80 to 120\% of their original size (independently per axis).

$\bullet$ \textbf{Translation:} we randomly crop or pad up to 25\% portion of the image with a probability of 50\%.

$\bullet$ \textbf{Flip:} we  randomly flip with a probability of 50\% (independently in horizontal and vertical directions).

\subsection{Performance}

Table \ref{tab:cub-results} shows the performance comparison of our proposed LCHP and other state-of-the-art methods on CUB-200-2011 testing dataset and ImageNet-1k validation dataset.

On CUB-200-2011 testing dataset,
our baseline (Pseudo Label) achieves 61.03\% \textit{Top-1} localization accuracy,
which is a strong performance compared with other state-of-the-art methods.
By training an additional regression network with pseudo-supervised learning, LCHP-I achieves 73.61\% \textit{Top-1} localization accuracy,
which outperforms our baseline model with 12.58\% accuracy.
The excellent performance of the LCHP-I model shows that the cross-image location information extraction in DDT is not necessary for PSOL.
Further, with learning consistency from high-quality pseudo-labels, our LCHP-R model achieves 80.39\%  \textit{Top-1} localization accuracy,
which outperforms our LCHP-I model with 6.78\% accuracy.
Compared to PSOL \cite{zhang2020rethinking}, our proposed LCHP achieves 5.42\% and 1.82\% improvement on \textit{Top-1} and \textit{GT-Konwn}  localization accuracy, respectively.

On ImageNet-1k validation dataset,
our LCHP-R achieves 54.12\% and 66.08\% accuracy on \textit{Top-1} and \textit{GT-Konwn} performances localization accuracy,
which outperforms our baseline with 6.91\% and 8.51\%, respectively.
Compared to other state-of-the-art methods, our proposed LCHP achieves comparable localization performance.

\begin{table}[ht]
	\caption{The performances (\%) comparison with state-of-the-art methods on Tiny-ImageNet validation dataset. The best performance has been bolded.}
	\label{tab:tiny-results}
	\centering
	\begin{tabular}{lccc}
		\toprule
		Method       				&  \textit{Top-1}  	& \textit{GT-Konwn}  			\\
		\midrule
		GR \cite{choe2018improved}				 		& 36.00 		& 57.82      				\\
		InfoCAM \cite{qin2019rethinking}				 		& 43.34 		& 57.79      				\\
		\midrule
		Pseudo Label  	 	& 41.24		 			& 50.33      				\\
		LCHP-I 				 	& 49.06	 				& 59.80      				\\
		LCHP-R		 		& \textbf{50.95}		 			& \textbf{61.87}      				\\
		\bottomrule
	\end{tabular}
\end{table}

Moreover, Table \ref{tab:tiny-results} shows the performance comparison on Tiny-ImageNet validation dataset.
Our proposed LCHP achieves 50.95\% and 61.87\% \textit{Top-1} and \textit{GT-Konwn} localization accuracy.
Compared to  InfoCAM \cite{qin2019rethinking}, our approach achieves 7.61\%  and  4.08\% improvement on \textit{Top-1} and \textit{GT-Konwn} localization accuracy, respectively.

\subsection{Ablation Study}

To further understand our approach, we design ablation experiments to study the effects of various parts of the network on the localization performance.

\begin{table}[ht]
	\caption{The \textit{GT-Konwn} localization  performance (\%) under different $\delta$ on CUB-200-2011  testing dataset. The best performance has been bolded.}
	\label{tab:ablation_delta}
	\centering
	\begin{tabular}{lccc}
		\toprule
		Method     		 	&  Pseudo Label 	& LCHP-I 			& LCHP-R				\\
		\midrule
		$\delta = 0.5$     	& 78.89				& 81.69 			& 88.63				\\
		$\delta = 0.6$      & \textbf{80.15}	& \textbf{88.95} 	& 93.09					\\
		$\delta = 0.7$      & 72.44				& 87.12				& \textbf{94.36}		\\
		$\delta = 0.8$      & 51.40				& 64.64				& 89.96				\\
		\bottomrule
\end{tabular}
\end{table}

Table \ref{tab:ablation_delta} shows the \textit{GT-Konwn} localization  performance under different $\delta$ on CUB-200-2011  testing dataset.
Experimental results show that our LCHP-R could achieve excellent performance improvements over the corresponding LCHP-I under different $\delta$.
It is worth noting that both Pseudo Label and LCHP-I achieve the best performance under $\delta = 0.6$, while LCHP-R achieves the best performance under $\delta = 0.7$.
Intuitively, the higher the localization accuracy of LCHP-I, the better the performance of LCHP-R.
We believe that the reason for this unusual phenomenon is mainly due to the definition of high-quality for pseudo bounding boxes.
Since the classification network is  trained on the raw image, the predicted bounding box that completely covers the image usually has a large enough $max(\hat{y}_{i,b}^{cls})$, 
and is then defined as a high-quality pseudo-label.
Therefore, this definition of high-quality will further cause the localization network to tend to predict larger bounding boxes in the refinement stage.
As a result, a slightly larger $\delta$ will offset this impact, so as to obtain better localization performance.

\begin{table}[htbp]
	\caption{The ablation study on the confidence threshold $\tau$ on CUB-200-2011 dataset. The best performance has been bolded.}
	\label{tab:ablation_tau}%
	\centering
	  \begin{tabular}{lccc}
	  \toprule
	  \multicolumn{1}{l}{\multirow{2}[1]{*}{Method}} & \multicolumn{2}{c}{Training set (LCHP-I)} & \multicolumn{1}{c}{Testing set (LCHP-R)} \\
			& \multicolumn{1}{c}{\textit{Nums}} & \multicolumn{1}{c}{\textit{GT-Known}} & \multicolumn{1}{c}{\textit{GT-Known}} \\
		\midrule
		$\tau = 0.0$ 			& 5994			& 84.01      				& \textit{failed} \\
		$\tau = 0.5$			& 4950 			& 85.89      				& 88.36	 \\
		$\tau = 0.6$			& 4519 			& 86.47	      				& 91.18 \\
		$\tau = 0.7$			& 3945 			& 87.52       				& 92.80 \\
		$\tau = 0.8$			& 3126			& 88.51	      				& 93.98 \\
		$\tau = 0.9$			& 1778			& \textbf{90.55}      		& \textbf{94.36} \\
	  \bottomrule
	  \end{tabular}%
  \end{table}%

To verify the effectiveness of the definition of high-quality, 
we experiment the localization performance of the high-quality  bounding boxes  predicted by LCHP-I on the CUB-200-2011 training dataset under different $\tau$,
which is shown in the left of Table \ref{tab:ablation_tau}. 
It is worth noting that this experiment is only to verify the effectiveness of the definition of high-quality.
We do not utilize any ground-truth bounding boxes in any training of our models.
\textit{Nums} means the number of high-quality pseudo-labels in the training dataset.
\textit{GT-Konwn} in the left means the \textit{GT-Konwn} localization accuracy of high-quality pseudo-labels in the training dataset.
When $\tau = 0$, all predicted bounding boxes are defined as high-quality pseudo-labels. 
The experimental results show that as $\tau$ increases, the localization performance of the pseudo-labels defined as high-quality is better.
which proves the validity of our definition of high-quality for bounding boxes.

We show the localization  performance of LCHP-R under different $\tau$ on CUB-200-2011  testing dataset on the right of Table \ref{tab:ablation_tau}.
It is observed that LCHP-R failed when $\tau = 0$, which shows that learning consistency from pseudo-labels with large deviations will decrease the localization performance.
Furthermore, in the case of $\tau \geq 0.5$, our LCHP-R can consistently outperform LCHP-I (\textit{GT-Konwn} localization accuracy is 87.12\%).
Moreover, with the increase of $\tau$ increases, the localization performance of our LCHP-R increases, which is consistent with the localization performance  of LCHP-I on the training set.

\begin{table}[ht]
	\caption{The \textit{GT-Konwn} localization performance (\%) under different backbone of the localization network on CUB-200-2011  testing dataset. The best performance has been bolded.}
	\label{tab:ablation_locbackbone}
	\centering
	\begin{tabular}{lcc}
		\toprule
		Backbone    			&  LCHP-I			&  LCHP-R	\\
		\midrule
		VGG19-BN    			&   80.19			&   83.95			\\
		ResNet50     			&   85.90			&   88.92			\\
		InceptionV3    			&    87.12 		&   \textbf{94.36} 			\\
		\bottomrule
\end{tabular}
\end{table}

Table \ref{tab:ablation_locbackbone} shows the \textit{GT-Konwn} localization performance under different backbone of the localization network on CUB-200-2011  testing dataset.
Experimental results show that our proposed LCHP-R outperforms the corresponding LCHP-I with at least 2.76\% on \textit{GT-Konwn} localization  performance, 
which proves that our LCHP methods are robust on different backbones of the localization network.

\begin{table}[ht]
	\caption{The localization  performance (\%) under different strategies of the strong augmentation on CUB-200-2011  testing dataset. The best performance has been bolded.}
	\label{tab:ablation_augmentation}
	\centering
	\begin{tabular}{cccc}
		\toprule
		Scale    			& Translation 	& Flip 			&  	\textit{GT-Konwn}			\\
		\midrule
		\checkmark  		&   			&  				&  	91.84			\\
		    				& \checkmark 	&  				&  	93.19			\\
		    				&   			& \checkmark 	&  	91.77			\\
		\checkmark    		& \checkmark  	&  				&  	93.80			\\
		\checkmark    		&   			& \checkmark 	&  	93.02			\\
		    				& \checkmark    & \checkmark 	&  	93.92			\\
		\checkmark  		& \checkmark  	& \checkmark 	&  	\textbf{94.36}			\\
		\bottomrule
\end{tabular}
\end{table}

Table \ref{tab:ablation_augmentation} shows
the localization  performance  on CUB-200-2011  testing dataset under different strategies of the strong augmentation.
When the three augmentation strategies are applied independently, 
they can all achieve positive localization performance improvements with at least 4.65\%  \textit{GT-Konwn} localization accuracy.
When the three aug strategies are applied together, the localization performance achieves the best localization performance.

\begin{figure}[h]
	\centering
	\includegraphics[scale=0.7]{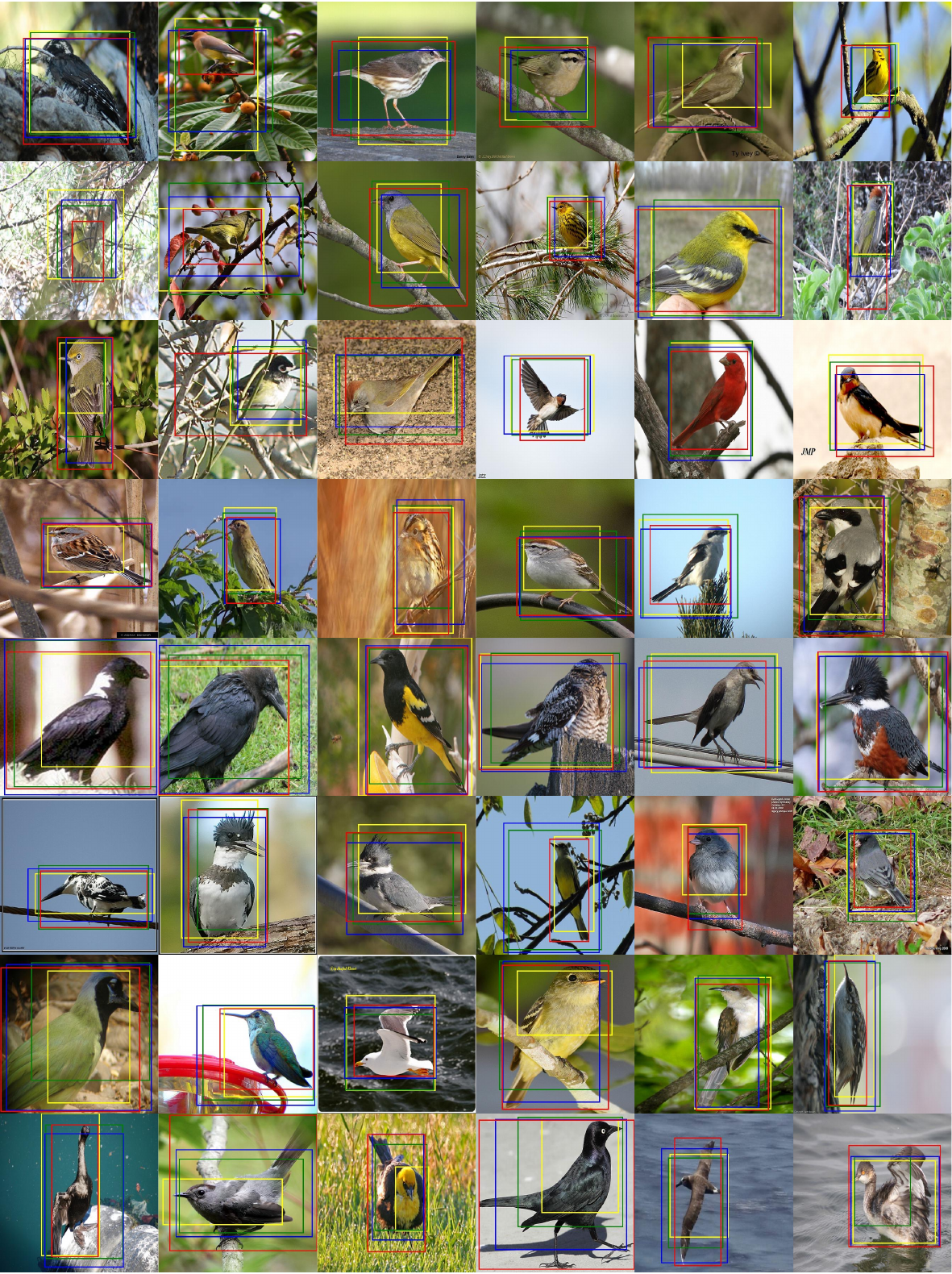}
	\caption{
		The visualization of predicted localization on randomly selected samples from  CUB-200-2011 testing dataset.
		The yellow bounding boxes are the prediction of our Pseudo Label method, 
		the green bounding boxes are the prediction of LCHP-I, 
		the blue bounding boxes are the prediction of LCHP-R, 
		and the red bounding boxes are the ground-truth bounding boxes.
	}
	\label{fig:vis_cub}
\end{figure}

\begin{figure}[h]
	\centering
	\includegraphics[scale=0.7]{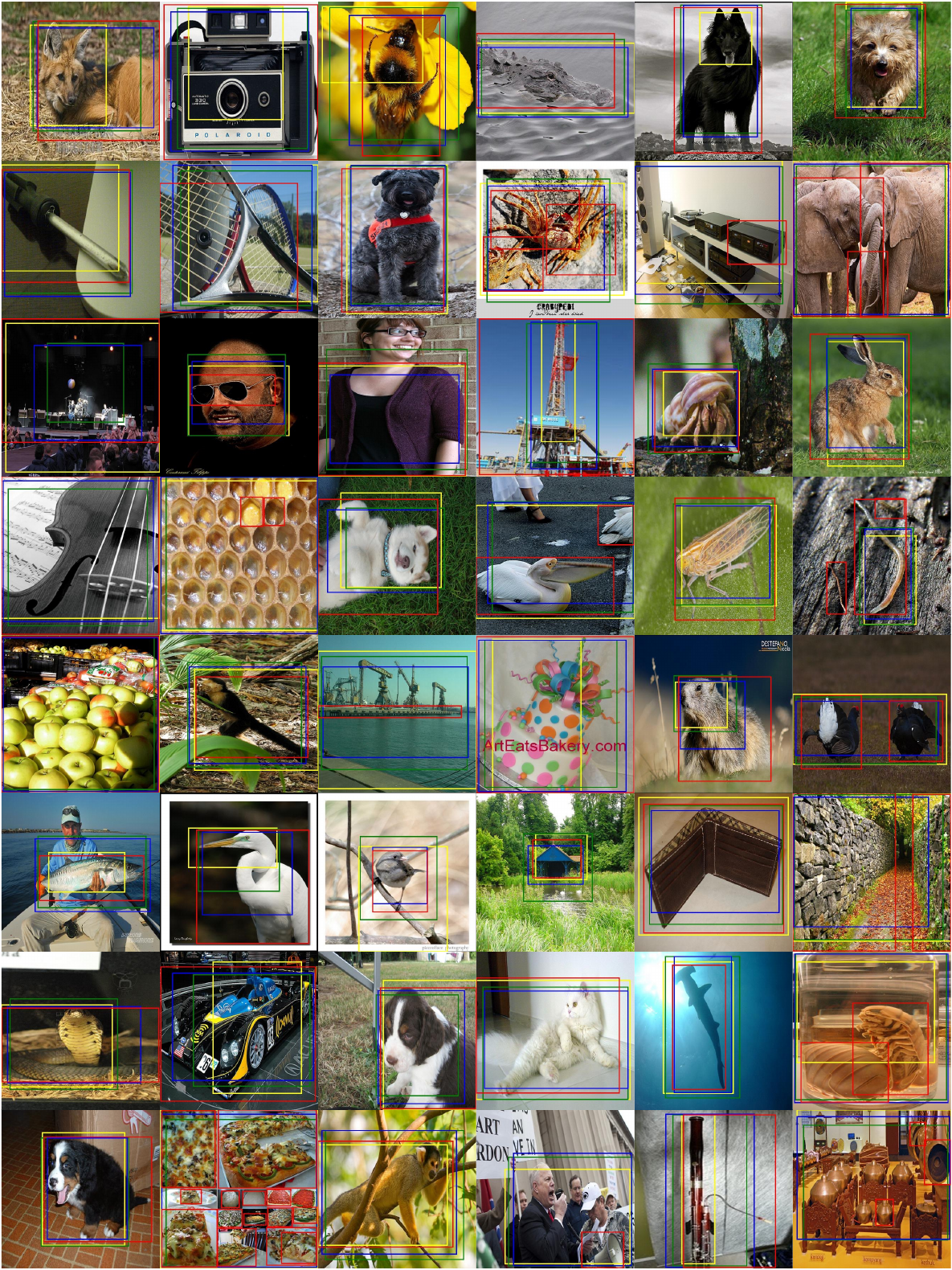}
	\caption{
		The visualization of predicted localization on randomly selected samples from  ImageNet-1k validation dataset.
		The yellow bounding boxes are the prediction of our Pseudo Label method, 
		the green bounding boxes are the prediction of LCHP-I, 
		the blue bounding boxes are the prediction of LCHP-R, 
		and the red bounding boxes are the ground-truth bounding boxes, which may be more than one red bounding box per image.
	}
	\label{fig:vis_imagenet}
\end{figure}

To evaluate the performance of our proposed LCHP intuitively, 
we visualized the predicted localization on randomly selected samples from CUB-200-2011 testing dataset and ImageNet-1k validation dataset,
which are shown in Fig.\ref{fig:vis_cub} and Fig.\ref{fig:vis_imagenet}, respectively.

\section{Conclusion}
In this paper, we propose a novel two-stage approach for weakly supervised object localization.
A simple and effective mask-based pseudo bounding box generator is proposed to generate high-precision bounding boxes for pseudo-labeling.
To refine the localization performance with consistency regularization, we propose a confidence evaluation method for retaining high-quality pseudo bounding boxes.
Our proposed approach  achieves state-of-the-art performance on CUB-200-2011 and Tiny-ImageNet, and achieves a comparable  performance compared with state-of-the-art methods on ImageNet-1k.

However, our LCHP relies on the assumption that there is only one instance in the image, which makes LCHP not perform well on the ImageNet-1k validation dataset.
Improving the performance of LCHP on multi-instance images could be our future work.

\section*{Acknowledgments}
This work is supported by Shanghai Frontier Science Research
Center for Gravitational Wave Detection.









\bibliographystyle{named}
\bibliography{ijcai22}

\end{document}